%% file: main.tex
\documentclass[10pt,twocolumn,letterpaper]{article}

\usepackage{cvpr}              

\usepackage{amsmath}
\usepackage{amssymb}
\usepackage{bm}
\usepackage{multirow}
\usepackage{graphicx}
\usepackage{caption}
\usepackage{subcaption}
\usepackage{algorithm}
\usepackage{algpseudocode} 
\usepackage{url}
\usepackage{xcolor}

\definecolor{cvprblue}{rgb}{0.21,0.49,0.74}
\usepackage[pagebackref,breaklinks,colorlinks,allcolors=cvprblue]{hyperref}

\def\paperID{3431} 
\def\confName{CVPR}
\def\confYear{2026}

\title{Fast-ARDiff: An Entropy-informed Acceleration Framework for Continuous Space Autoregressive Generation}
\author{Zhen Zou$^{1}$ \quad Xiaoxiao Ma$^{1}$ \quad Jie Huang$^{1,2}$ \quad Zichao Yu$^{1}$ \quad Feng Zhao$^{1}$\\
\normalsize
$^{1}$University of Science and Technology of China \quad
$^{2}$JD Joy future AI\\
{\tt\small zouzhen@mail.ustc.edu.cn}
}

\begin{document}
\maketitle
\input{sec/0_abstract}    
\input{sec/1_intro}
{
    \small
    \bibliographystyle{ieeenat_fullname}
    \bibliography{main}
}

\input{sec/X_suppl}
\end{document}

%% file: sec/0_abstract.tex
\begin{abstract}

Autoregressive(AR)-diffusion hybrid paradigms combine AR’s structured modeling with diffusion’s photorealistic synthesis, yet suffer from high latency due to sequential AR generation and iterative denoising.
In this work, we tackle this bottleneck and propose a unified AR–diffusion framework Fast-ARDiff that jointly optimizes both components, accelerating AR speculative decoding while simultaneously facilitating faster diffusion decoding.
Specifically: (1) The entropy-informed speculative strategy encourages draft model to produce higher-entropy representations aligned with target model’s entropy characteristics, mitigating entropy mismatch and high rejection rates caused by draft overconfidence.
(2) For diffusion decoding, rather than treating it as an independent module, we integrate it into the same end-to-end framework using a dynamic scheduler that prioritizes AR optimization to guide the diffusion part in further steps. 
The diffusion part is optimized through a joint distillation framework combining trajectory and distribution matching, ensuring stable training and high-quality synthesis with extremely few steps.
During inference, shallow feature entropy from AR module is used to pre-filter low-entropy drafts, avoiding redundant computation and improving latency.
Fast-ARDiff achieves state-of-the-art acceleration across diverse models:
on ImageNet 256$\times$256, TransDiff attains 4.3$\times$ lossless speedup, and NextStep-1 achieves 3$\times$ acceleration on text-conditioned generation. Code will be available at \textcolor{cyan}{\url{https://github.com/aSleepyTree/Fast-ARDiff}}.

\end{abstract}

%% file: sec/1_intro.tex
\section{Introduction}
\label{sec:intro}

\begin{figure}
    \centering
    \includegraphics[width=1\linewidth]{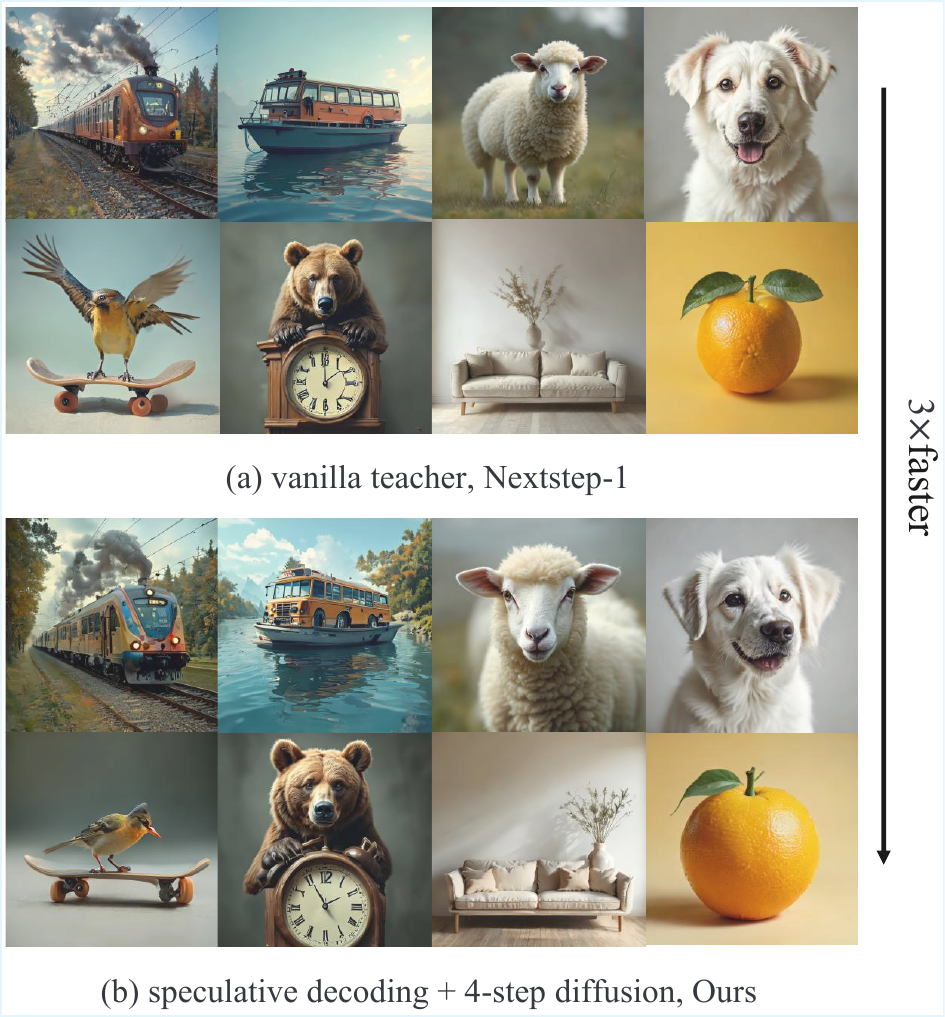}
    \caption{Qualitative generation comparison between the vanilla NextStep-1~\cite{team2025nextstep} and our method on GenEval~\cite{ghosh2023geneval}.
    }
    \label{fig1:show}
\end{figure}

AR models have witnessed great success in natural language processing, with large language models (LLMs) ~\cite{achiam2023gpt4,anil2023palm,touvron2023llama} demonstrating strong abilities in reasoning and open-ended generation. Motivated by this success, recent studies extend AR modeling to visual generation~\cite{liu2024lumina_mgpt,sun2024llamagen}, leveraging its strength in capturing long-range dependencies and structured semantics.
Early AR models generated pixels directly~\cite{van2016pixelcnn}, but suffered from low resolution.
With advances in vector quantization (VQ)~\cite{van2017vqvae, esser2021vqgan}, generation shifted to discrete token space, enabling high-level feature modeling and higher resolutions.
Recent works~\cite{tian2024var,wang2024par} further introduced parallel decoding for higher efficiency.

However, AR models are constrained by quantization errors that degrade fine-grained details and limit generation fidelity.
Recent works~\cite{razzhigaev2023movq,yu2023lfq,zhao2024bsq} attempt to improve quantizers through grouped codebooks or predefined quantization rules to stabilize training, yet they still rely on discrete representations and thus incur inevitable information loss.
Another line of work, hybrid AR-diffusion paradigms~\cite{zhen2025marrying,ren2024flowar,team2025nextstep} have emerged as promising solutions, which shift AR generation to continuous latent spaces and avoid quantization altogether. 
MAR~\cite{li2024mar} uses diffusion-inspired denoising losses on continuous latents, and FlowAR~\cite{ren2024flowar} combine scale-wise AR with flow matching in continuous space to preserve semantics. 
Furthermore, this hybrid architecture leverages AR’s structured semantic modeling to guide the diffusion module while exploiting diffusion’s high-fidelity synthesis, effectively balancing the trade-offs between the two paradigms.
Nevertheless, the two-stage nature of these hybrids can introduce efficiency challenges, as sequential AR computation and iterative diffusion denoising jointly cause high latency.

\begin{figure*}
    \centering
    \includegraphics[width=1\linewidth]{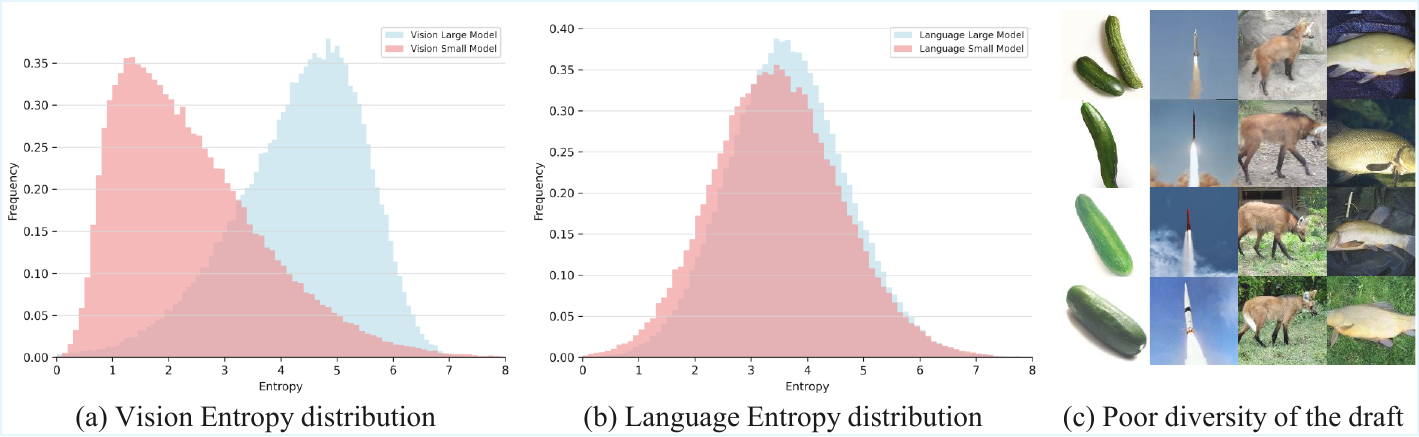}
    \caption{Entropy distribution comparison between vision and language models, and poor diversity of small draft models. (a) Entropy distribution of vision models (small vs large AR models); (b) Entropy distribution of language models; (c) Insufficient diversity of outputs from small draft models, which generate similar images .
    }
    \label{fig2:moti}
\end{figure*}

Few studies have explored acceleration for this hybrid paradigm, and simply decoupling the AR and diffusion parts for separate acceleration proves insufficient. A straightforward solution is to combine speculative decoding~\cite{li2024eagle,li2024eagle2} for the AR component with distillation-based acceleration~\cite{yin2024one,yin2024improved} for the diffusion part within an end-to-end framework.
However, we find that this leads to an excessively high rejection rate of small models, reducing the efficiency of speculative decoding and thus limiting the performance of diffusion distillation.
As illustrated in Figure~\ref{fig2:moti} (a), the root cause lies in an \textbf{entropy mismatch between the small and large models}: the small draft produces features with extremely low entropy, indicating overconfident and homogeneous predictions, while the large verifier maintains a far more uniform entropy distribution.
We attribute it to the inherently low information density of images—unlike text, where each token carries distinct semantics and inherently preserves diversity, images often contain redundant regions (e.g., skies, blank walls) with minimal variation.
As shown in Figure~\ref{fig2:moti} (a)(b), language models remain consistent distributions across scales, whereas vision models exhibit drastic entropy shifts, which we attribute to the overfitting on redundant regions.
This consistency explains why speculative decoding excels in language processing and further highlights the unique challenges of adapting it to image-based AR+Diffusion frameworks.

To address the above challenges, this paper proposes a unified acceleration framework Fast-ARDiff for AR+Diffusion hybrids, with three core designs: 
First, we design an entropy-informed speculative decoding strategy to alleviate the entropy mismatch between small and large AR models, reducing the rejection rate of draft models.
Our approach introduces a feature-level entropy loss that explicitly encourages the draft model to produce higher-entropy, more diverse representations.
In addition, we develop a joint diffusion distillation framework combining trajectory distillation (for stable initialization) and distribution matching (for extreme step reduction), mitigating the latency of iterative denoising.
Finally, we unify the training of speculative AR and diffusion distillation in an end-to-end manner to ensure cross-module synergy. A dynamic loss weighting scheme linearly anneals the focus from AR alignment to diffusion distillation during training, prioritizing semantic guidance early and refining visual synthesis later. 
Our main contributions can be summarized as follows:
\begin{itemize}
    \item We identify that the entropy mismatch between AR models with different sizes in the AR+Diffusion paradigm leads to a high rejection rate for speculative decoding.
    \item We propose an end-to-end unified framework that jointly optimizes both AR and diffusion via entropy-informed speculative decoding and two-stage diffusion distillation.
    \item Extensive experiments on multiple AR+Diffusion hybrids demonstrate that our method achieves SOTA performance with minimal loss in visual quality and diversity.
\end{itemize}

\section{Related Work}

\subsection{ Diffusion and AR Models for Generation}

Diffusion and AR remain dominant in high-fidelity generative modeling, with recent progress focusing on scalability and efficiency~\cite{peebles2023scalable,qin2024worldsimbench,seedream2025seedream,chang2022maskgit,sun2024autoregressive,tian2024var,razavi2019generating}.

For diffusion models, advances have focused on scaling to high resolution, accelerating sampling, and expanding to video. 
DiT~\cite{peebles2023scalable} is the first work to use transformers instead of U-Nets and to explore the scalability of diffusion transformers.
Seedream 4.0~\cite{seedream2025seedream}, an advanced multimodal image generation framework that integrates an efficient and scalable diffusion transformer with a high-compression VAE, achieving more than ten times the  acceleration of the previous model, delivering superior performance in all aspects.

For AR models: MaskGit~\cite{chang2022maskgit} revolutionized AR efficiency with ``masked generation": instead of generating tokens sequentially, it iteratively predicts and refines masked token groups, reducing inference steps by 10× while retaining FID 1.8 on ImageNet 256×256. Pure AR models also advanced: LlamaGen~\cite{sun2024autoregressive}, a 7B-parameter language-aligned AR model, treats images as ``visual sentences" and uses a transformer with 2D positional encoding to generate pixel values directly.

\subsection{AR+Diffusion Hybrid Paradigm for Generation}
The AR+Diffusion hybrid bridges AR’s structured semantic modeling and diffusion’s high-fidelity synthesis, addressing pure AR’s latency and pure diffusion’s semantic misalignment .
Early explorations laid the groundwork by validating the feasibility of integrating AR’s sequential logic with diffusion-like probabilistic modeling. MAR~\cite{li2024mar} blurred the lines between AR and diffusion: while retaining AR’s core of predicting masked tokens in sequence, it introduced a diffusion-inspired denoising loss to refine predictions. By operating on continuous latent features (avoiding discrete vector quantization), MAR eliminated the loss of information from VQ-based AR models and demonstrated that diffusion-style optimization could enhance AR’s generation quality, a significant leap over prior VQ-AR models.
TransDiff~\cite{zhen2025marrying} advanced to end-to-end integration: an AR transformer encoded semantic features, and a diffusion decoder used them for guidance, resolving AR’s \(O(n^2)\)complexity and diffusion’s ambiguity. FlowAR~\cite{ren2024flowar} replaced diffusion with flow matching, adding ``Spatial-adaLN" for feature alignment . NextStep-1~\cite{team2025nextstep} unified AR and flow matching in one transformer, reducing generation latency.
Collectively, these models define the AR+Diffusion paradigm’s core principles: (1) AR provides hierarchical, sequential semantic guidance (either via discrete tokens or continuous features); (2) diffusion/flow matching handles high-dimensional visual synthesis to ensure photorealism; (3) tight integration (via joint training or adaptive feature injection) is critical to preserving alignment between guidance and synthesis. While early works like MAR focused on loss-level fusion, recent models like TransDiff and FlowAR have advanced to architectural co-design, setting the stage for more efficient and controllable generative systems.

\subsection{Acceleration of Generative Models}

Acceleration techniques for generative models have matured into two distinct streams, tailored to the unique bottlenecks of AR and diffusion models~\cite{tian2024var,wang2024par,bolya2023token,bolya2022token,lou2024token,yuan2024ditfastattn,Structuralpruning,zhang2024laptop,shang2023post,so2024temporal}. While AR models struggle with sequential computation latency, diffusion models are limited by iterative denoising steps—each demanding specialized optimizations. However, synergistic acceleration for AR+Diffusion hybrids remains largely unaddressed, with current methods optimizing each component in isolation.

The core bottleneck of AR models lies in their sequential token generation, where each step depends on all prior outputs, leading to latency for $n$ tokens. Key acceleration strategies target reducing redundant computations and compressing model size. Diffusion models’ primary inefficiency stems from iterative denoising and large U-Net/Transformer backbones. Acceleration strategies focus on reducing steps, compressing models, and reusing computations. Distribution Matching Distillation (DMD)~\cite{yin2024one} advanced this by aligning student and teacher distributions over the entire latent space (not just individual steps), enabling 100× speedup with FID 2.62 on ImageNet 64×64. Its successor, DMD2~\cite{yin2024improved}, added adversarial loss to eliminate regression loss, achieving single-step generation with FID 1.28 on ImageNet 64×64. For large models like SDXL~\cite{podell2023sdxl}, SDXL-Lightning~\cite{lin2024sdxl} combined progressive distillation with DiffusionGAN loss, reducing steps to 4 while maintaining photorealism in 1024×1024 images.

\input{sec/3_preliminary_xiaoxiao}

\section{Method}
\begin{figure*}
    \centering
    \includegraphics[width=0.85\linewidth]{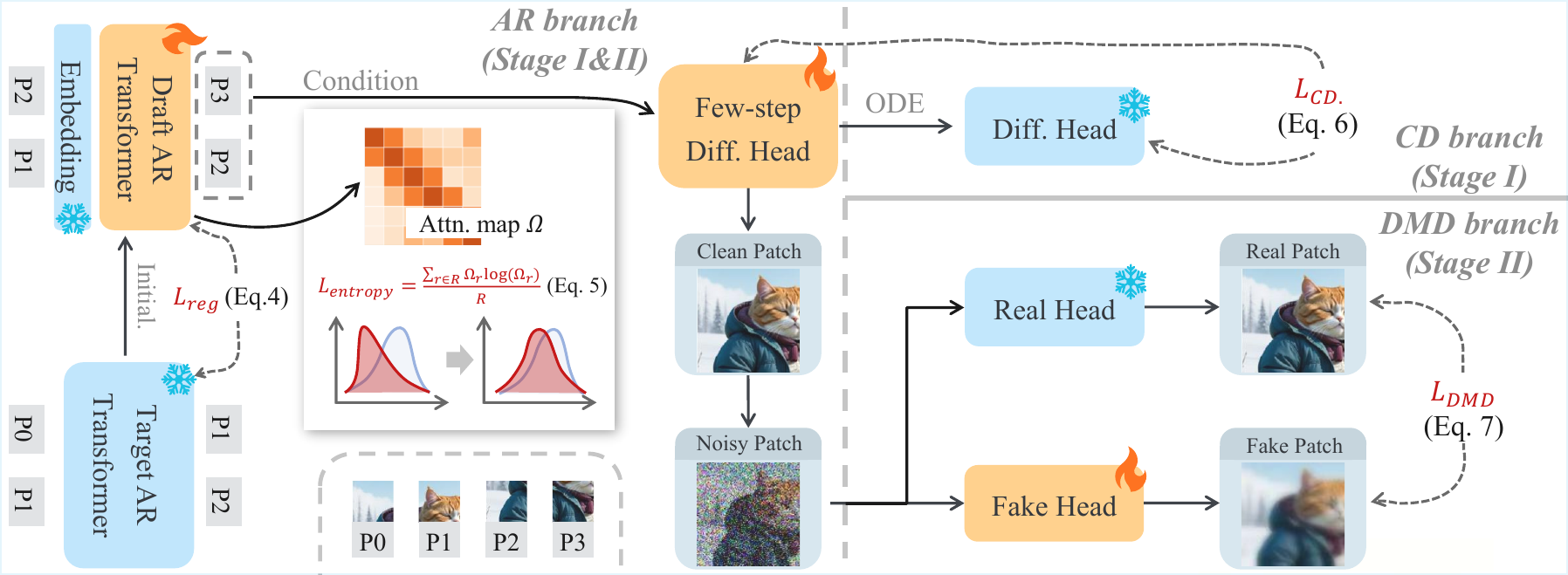}
    \caption{
    Illustration of Fast-ARDiff: an end-to-end unified framework enabling mutual perception between speculative decoding (AR branch) and diffusion distillation (CD/DMD branches). (Left) The entropy-informed loss alleviates entropy mismatch in the AR draft model for better speculative generation, while providing entropy-aligned semantic guidance to stabilize subsequent diffusion distillation. (Right) Two-stage diffusion distillation feeds back into the end-to-end pipeline, dynamically regulating AR’s feature learning to align with diffusion’s high-fidelity demands. This bidirectional interaction(entropy guidance refining diffusion initialization and diffusion loss optimizing AR modeling) enables tight cross-module synergy via end-to-end optimization.
    }
    \label{fig3:method}
\end{figure*}

As shown in Figure~\ref{fig3:method}, Fast-ARDiff consists of three core components: entropy-informed speculative decoding, step distillation with initialization adaptation, and a unified training-inference pipeline. These components are jointly designed to address the unique challenges of AR+Diffusion acceleration, including entropy mismatch in speculative decoding and training instability in diffusion distillation. 

\subsection{Entropy-Informed Speculative Decoding}
Speculative decoding is a natural candidate for accelerating the AR component of AR+Diffusion models, but direct application of vanilla frameworks (e.g., EAGLE) yields poor performance. Through empirical analysis, we identify two critical issues:
\begin{itemize}
 \item \textbf{Low hit rate of the small draft model:} The draft model’s candidate outputs are frequently rejected by the large target model, leading to repeated target evaluations and consequently failing to reduce latency.
 \item \textbf{Degraded sample diversity:} As illustrated in Figure~\ref{fig2:moti} (c), the small model’s low-entropy outputs result in diminished diversity in generated samples, further exacerbating the gap between draft and target model behaviors.
\end{itemize}
Further analysis indicates that these issues are due to the entropy mismatch between the draft model and the target model. As shown in Figure~\ref{fig2:moti} (a), the small model exhibits a significantly lower entropy in its output compared to the large model, indicating overly confident predictions with insufficient diversity. This mismatch directly leads to poor performance. 

To resolve these issues, we propose entropy-informed speculative decoding, building on EAGLE’s paradigm while introducing critical modifications:
\begin{itemize}
 \item \textbf{Eliminate token-level cross-entropy loss:} Unlike EAGLE, which optimizes both feature and token alignment, we remove token-level loss $\mathcal{L}_{\text{cls}}$ in Eq.~\ref{eq_loss_reg_cls} to keep consistent with the continuous-space nature of AR+Diffusion.
 \item \textbf{Retain feature-level regression loss:} We preserve the $L1$ regression loss between the draft model’s predicted features and the target model’s true features, ensuring structural alignment of semantic guidance.

 \item \textbf{Add feature-level entropy loss:} As shown in Figure~\ref{fig3:method}, a novel entropy loss is introduced to explicitly encourage higher entropy in the draft model’s outputs, Let $\Omega_r$ denote the $r-th$ row of the attention map from the penultimate layer of transformer body after softmax:
 \begin{equation}
     \mathcal{L}_{\text{entropy}}=\frac{\Sigma_{r \in R} \Omega_r log(\Omega_r)}{R},
    \label{eq_entropyloss}
 \end{equation}
where $\Omega_r$ denotes the $r$-th row after softmax normalization of the second-to-last attention map of the AR module.
This loss penalizes low-entropy feature distributions, forcing the draft model to generate more diverse candidates that better match the target model’s entropy.
\end{itemize}

\subsection{Step Distillation with Initialization Adaptation}
\label{sec4.2}
DMD is a state-of-the-art technique for accelerating diffusion models by aligning student-teacher distributions. However, directly applying DMD to the diffusion component of AR+Diffusion hybrids causes immediate training collapse. We trace this instability to DMD’s strict requirement for high-quality initializations: the diffusion student model must already exhibit reasonable performance at low steps to effectively learn from the teacher. 

To address this, we propose a two-stage step distillation strategy: initialization adaptation followed by Distribution Matching Distillation. 
This design is grounded in our insight into Consistency Models, specifically, their ability to straighten the generated trajectory.
By capturing the continuous evolution law of the denoising process, this modeling eliminates distribution breaks between discrete steps through trajectory straightening. This aligns perfectly with DMD’s need for high-quality initialization: it ensures that the student model already has a trajectory basis consistent with the teacher’s distribution at low steps, while preventing optimization collapse caused by chaotic initial trajectories.

\noindent \textbf{Consistency Distillation for Diffusion Head Initialization:} We first perform consistency distillation on the diffusion head, which ensures the student model achieves baseline low-step performance, providing a viable initialization:
\begin{equation}
\mathcal{L}_{CD} = \mathop{\mathbb{E}}_{\mathop{x_0 \sim p_{data}}\limits_{t\sim \text{Uniform}(0,T)} } \left[ \left\| \text{Diff}_{\theta}(x_{t}) - \text{Diff}_{\theta}(x_{t-1}) \right\|_2^2 \right],
\end{equation}
where $x_{t-1}$ is obtained from $x_t$ by ODE. 
This process essentially enables the diffusion head to mimic the logic of Consistency Models in modeling the mean velocity over the interval $[0,1]$—by enforcing consistency in denoising outputs for the same input under different noise perturbations, it learns the continuous evolution law of the teacher model’s denoising trajectory rather than isolated step behaviors, laying an initialization foundation of alignment for DMD.

\noindent \textbf{DMD for Extremely Low Step Generation:} Building on the well-initialized student, we apply DMD to align the student’s latent distribution with the teacher’s, allowing further reduction in sampling steps without sacrificing quality:
\begin{equation}
    \begin{aligned}
        \mathcal{L}_{DMD} &= D_{\text{KL}}(p_{\text{fake}} \parallel p_{\text{real}}) = \mathop{\mathbb{E}}_{x \sim p_{\text{fake}}} \left( \log \frac{p_{\text{fake}}(x)}{p_{\text{real}}(x)} \right)\\
    &= \mathop{\mathbb{E}}_{\mathop{z \sim \mathcal{N}(0,I)}\limits_{ x=G_\theta(z)}} \left( - \left( \log p_{\text{real}}(x) - \log p_{\text{fake}}(x) \right) \right).
    \end{aligned}
\end{equation}
This two-stage approach leverages consistency distillation to meet DMD’s initialization requirements, avoiding training collapse. We randomly interpolate between $X_0$ and $X_T$ to construct the input instead of selecting $n$ fixed points($X_{\frac{T}{n}}$), which ensures the perception of diffusion models on variant noise levels.

\subsection{Unified Training and Inference Pipeline}

\noindent \textbf{Dynamic Loss Weighting in Training Phase:} Given the distinct roles of AR (semantic guidance) and diffusion (image synthesis) in the hybrid paradigm, we use a linearly annealed weight \(\alpha(t)\) for total loss to co-optimize AR and diffusion components, prioritizing AR alignment in early training and diffusion distillation later. The annealing function is:
\begin{equation}
\alpha(t) = \alpha_0 - (\alpha_0 - \alpha_T) \cdot \frac{t}{T_{\text{total}}},
\end{equation}
where \(\alpha_0=0.95\) (initial weight for AR losses), \(\alpha_T=0.05\) (final weight for AR losses),$t$ is the current training epoch. The total loss is:
\begin{equation}
    \mathcal{L}_{\text{total}} = \alpha(t) \cdot (\mathcal{L}_{reg} + \mathcal{L}_{entropy}) + (1 - \alpha(t)) \cdot \mathcal{L}_{dist},
\end{equation}
where $L_{dist}$ represents $\mathcal{L}_{CD}$ or $\mathcal{L}_{DMD}$ in Section.~\ref{sec4.2}. $\mathcal{L}_{reg}$ is defined in Eq.~\ref{eq_loss_reg_cls} and $\mathcal{L}_{entropy}$ is defined in Eq.~\ref{eq_entropyloss}.

\noindent \textbf{Entropy-Based Early Stopping in Inference Phase:} 
We follow the speculative decoding strategy in the continuous space proposed in  \cite{wang2024continuous}. 
The draft model sequentially generates $n$ tokens, which are then validated in parallel by the target model to achieve acceleration. The core principle is to avoid low-entropy features.
But the reality is that even entropy-regulated draft models may produce low-entropy trajectories in certain scenarios. We find that for a specific generated sample, the entropy discrepancy across different layers of the model is negligible. This allows us to infer the diversity of the final output of the draft model  using entropy statistics from shallow-layer features.

To alleviate the impact of residual low-entropy behavior in the draft model, we introduce an early entropy monitoring mechanism. 
During speculative decoding, we track the entropy of attention maps in the shallow layers of the draft model. If low entropy is detected at this early stage (signaling that the draft model is generating overly confident, low-diversity features), we immediately terminate speculation and switch to the large target model for AR generation. The threshold is calculated from the training set’s entropy loss statistics:
\begin{equation}
\tau = 0.3 \cdot \mathbb{E}_{\mathcal{D}_{\text{train}}}\left[ -\mathcal{E}_{\text{entropy}}^{l_s} \right] - 0.1 \cdot \text{std}\left( -\mathcal{E}_{\text{entropy}}^{l_s} \mid \mathcal{D}_{\text{train}} \right),
\end{equation}
where $\mathcal{E}_{\text{entropy}}^{s}$ denotes the shallow-layer entropy indicator, calculated with the same formulation as Eq.~\ref{eq_entropyloss} with feature from shallow layer $\Omega^s$ of the draft model.
This adaptive strategy ensures that even when the draft model fails to meet the entropy requirements, inference can still proceed with high-quality guidance from the target model, balancing inference speed and generation quality.

\section{Experiment}

\begin{figure}
    \centering
    \includegraphics[width=1\linewidth]{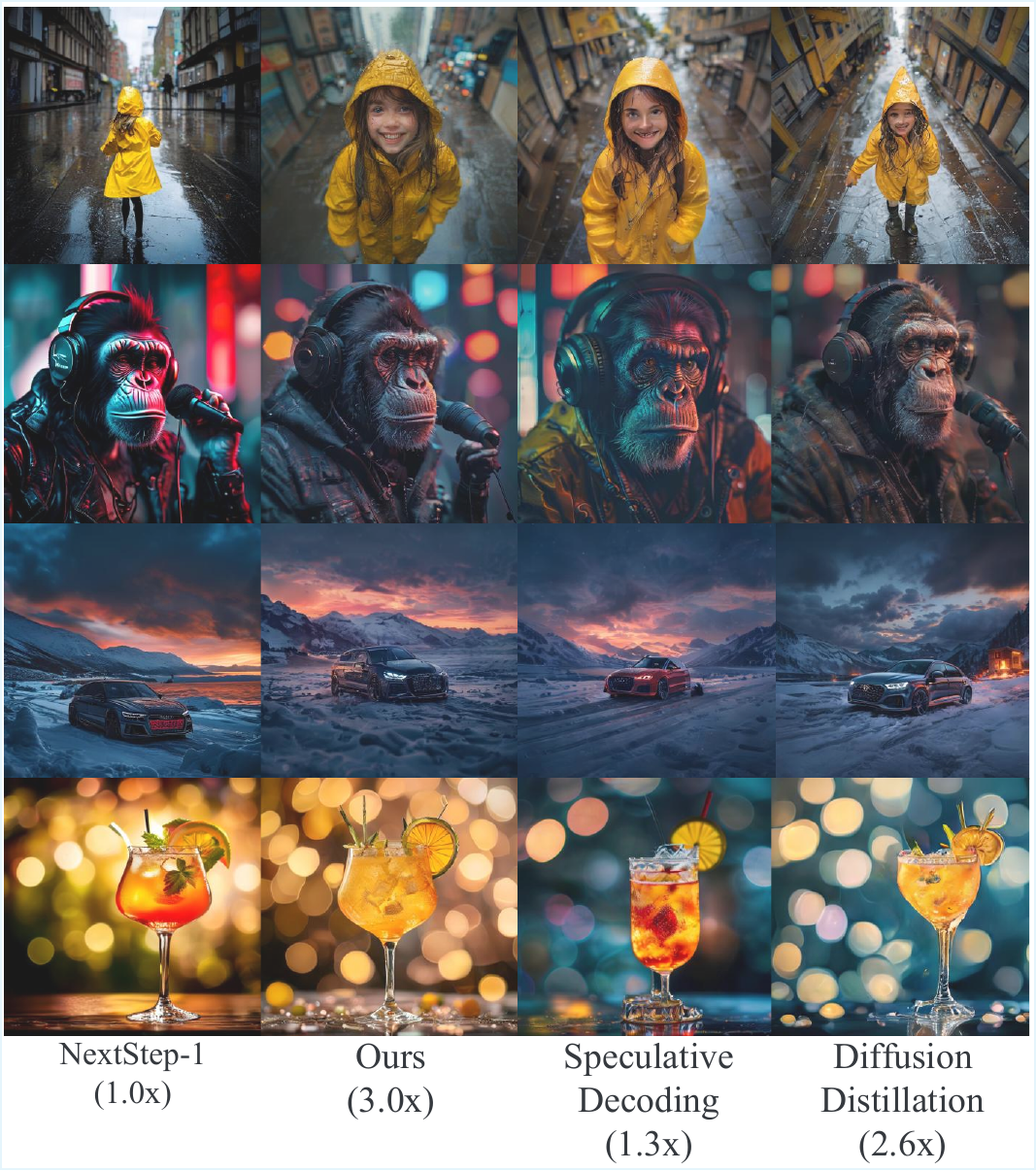}
    \caption{Comparison with NextStep-1~\cite{team2025nextstep} on  MJHQ-30K~\cite{li2024playground}.    }
    \label{t2i_show}
\end{figure}

\begin{figure*}[tp]
    \centering
    \includegraphics[width=1 \linewidth]{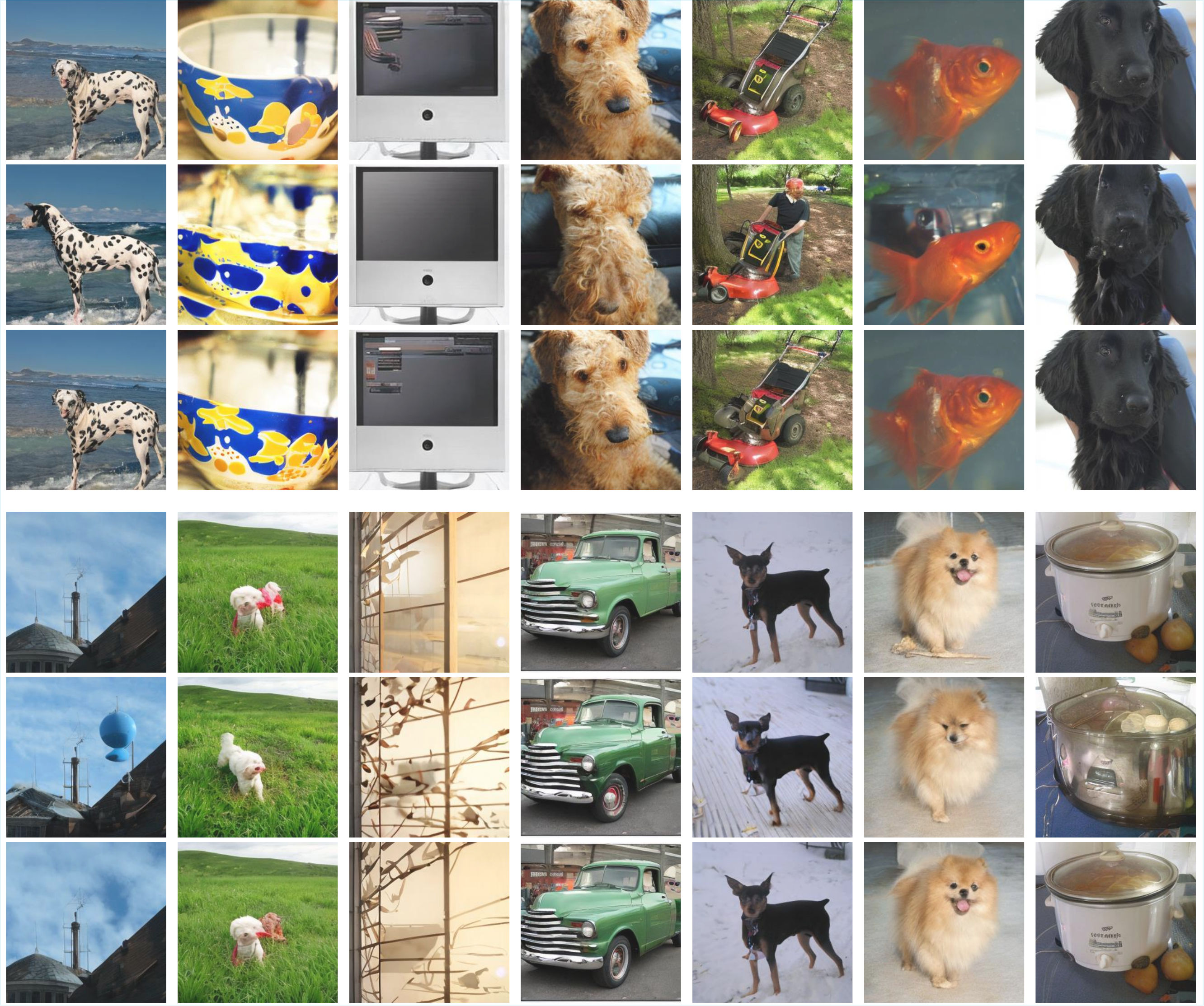}
    \caption{Comparison on ImageNet 256 $\times$ 256, TransDiff(Row 1), TransDiff+DMD(Row 2), TransDiff+Ours(Row 3).
    }
    \label{imagenet_show}
    \vspace{-1em}
\end{figure*}

\subsection{Impulment details}
To validate the effectiveness of our proposed acceleration framework for AR+Diffusion hybrids, we conduct extensive experiments on representative models. Given the lack of a unified architecture in AR+Diffusion paradigms—with AR components varying from mask-based generation, VAR-style generation, to vanilla next-token prediction (NTP), and diffusion components ranging from MLP-based to Transformer-based designs. We cover a broad range of architectures to ensure comprehensive validation, including those whose AR do not inherently support speculative sampling. Specifically, we select representative frameworks: MAR~\cite{li2024mar}, TransDiff~\cite{zhen2025marrying}, and NextStep-1~\cite{team2025nextstep}.

We compare our framework against the original implementations of TransDiff, MAR, and NextStep-1, as well as state-of-the-art acceleration methods for AR and Diffusion (where applicable), including vanilla speculative decoding and standalone DMD diffusion distillation. 
Take TransDiff-L for example, the draft model contains 6 transformer blocks initialized with the target. We train the model 20 epochs for stage 1(CD) and 10 epochs for stage 2(DMD), where AR related loss remains unchanged.
All models are evaluated according to the settings of the original papers. For the ImageNet experiments, we assess FID~\cite{heusel2017gans}, IS~\cite{salimans2016improved}, and latency on 50,000 images. For the text-to-image task, we train our model on the CC3M dataset~\cite{sharma2018conceptual} and  evaluate on GenEval~\cite{ghosh2023geneval} and MJHQ-30K~\cite{li2024playground}.

\begin{table}[h]
\centering
\caption{Class-conditional generation on ImageNet 256×256.}
\label{tab:imagenet_gen}
\resizebox{0.99\linewidth}{!}{
\begin{tabular}{lllll}
\toprule
\textbf{Method} & \textbf{Latency/s}& \textbf{Speedup}  & \textbf{FID ($\downarrow$)} & \textbf{IS ($\uparrow$)}\\
\midrule
\multicolumn{4}{l}{\textbf{\textit{MAR}}} \\

\midrule
MAR-L & 5.31 & 1.00 & 1.78 &296.0 \\
 + DMD&  1.99 & 2.67 & 1.81 &295.5 \\
 + LazyMAR&  2.29 & 2.32 & 1.93 &297.4 \\
 + Ours  & 1.21 & 4.38  & \textbf{1.77} &\textbf{297.1} \\
MAR-H & 9.97&1.00  & 1.55 & 303.7 \\
 + DMD& 4.17  & 2.39& 1.73 & 301.0 \\
 + LazyMAR& 4.24 &2.35 &  1.69 & 299.2 \\
 + Ours & 2.23 &4.47 &  \textbf{1.61} & \textbf{304.3} \\
\midrule
\multicolumn{4}{l}{\textbf{\textit{TransDiff}}} \\

\midrule
TransDiff-L & 3.17 & 1.00 & 1.61 & 295.1 \\
 + SD  &1.75 &1.81 & 1.88 & 283.6 \\
 + DMD  & 1.61&1.97 & 1.79 & 288.3 \\
 + Ours  &0.68 & {4.69} & \textbf{1.66} & \textbf{294.1} \\
TransDiff-H & 6.72 & 1.00 & 1.55 & 297.9 \\
 + SD & 3.52  & 1.91 & 1.68 & 291.1 \\
 + DMD & 3.48 & 1.93 & 1.71 &  289.9 \\
 + Ours & 1.38 & 4.88 & \textbf{1.60} & \textbf{295.0} \\
\bottomrule
\end{tabular}
}
\end{table}

\subsection{Comparision}
As detailed in Table.~\ref{tab:imagenet_gen} and \ref{tab:proprietary_gen}, our method demonstrates remarkable performance. For Class-conditional generation on ImageNet, Our method achieves more than 4$\times$ speed-up on both baseline. In the text-to-image task, our method is nearly 3$\times$ faster with even better performance. Qualitative examples are shown in Figure~\ref{t2i_show} and \ref{imagenet_show}. These results demonstrate its ability to achieve promising results across a wide range of generative tasks.
\subsection{Ablation Study}
To isolate the contribution of each component in our framework, we conduct extensive ablation experiments, systematically removing or modifying key modules, and evaluating their impact on generation quality, acceleration efficiency, and training stability. The ablation variants are defined as follows, with all other settings consistent with our full framework.

\noindent \textbf{$\mathbb{A}$: Ablating Entropy Loss.} We remove the feature-level entropy loss from the training objective to tests whether the entropy loss is critical for resolving the entropy mismatch between the draft and target AR models.

\noindent \textbf{$\mathbb{B}$: Ablating Stage-Wise Loss Reweighting.} Instead of using the dynamic annealing function \(\alpha(t)\) to balance losses, we fix the weights as \(\alpha = 0.5\) throughout the training. This variant examines whether adaptive reweighting is necessary to coordinate the training of AR (semantic guidance) and diffusion (image synthesis) components.

\noindent \textbf{$\mathbb{C}$: Ablating Consistency Distillation Initialization.} We skip the first stage of consistency distillation and directly apply DMD to the diffusion student model, starting from random initialization. This variant validates the necessity of our two-stage distillation strategy (consistency distillation → DMD) to avoid training collapse.

\noindent \textbf{$\mathbb{D}$: Ablating Entropy-Based Early Stopping in Inference.} We disable the early monitoring of attention map entropy during inference, allowing the draft model to generate candidates even when low entropy is detected. This variant assesses whether the early stopping mechanism is effective in mitigating residual low-quality draft output.

\begin{table}[h]
\centering
\caption{Comparison on GenEval and MJHQ-30K.}
\label{tab:proprietary_gen}
\resizebox{0.99\linewidth}{!}{
\begin{tabular}{llllll}
\toprule
\textbf{Method} & \textbf{Speedup} &\textbf{GenEval$\uparrow$} & \multicolumn{2}{c}{{\textbf{MJHQ-30K}}}  \\
& & &\textbf{FID ($\downarrow$)} & \textbf{CLIP ($\uparrow$)} & \\
\midrule
NextStep-1  & 1.00 & {0.63} & 6.71 &28.67 \\
 + SD & 2.03 & 0.63 &  6.90 & 27.96 \\
 + DMD  & 1.66 & 0.59 & 8.33 & 25.19  \\
 + Ours  & 2.97 & \textbf{0.66} & \textbf{6.68} & \textbf{28.88} \\
\bottomrule
\end{tabular}
}
\end{table}

\begin{table}[h]
\centering
\caption{Ablation study with TransDiff on ImageNet 256×256.}
\label{tab:Ablation}
\begin{tabular}{lrr}
\toprule
\textbf{Method}   & \textbf{FID ($\downarrow$)} & \textbf{IS ($\uparrow$)}\\

\midrule
TransDiff-H  & 1.55 & 297.9 \\
Ours w/o $\mathbb{A}$ &  1.72 & 289.3 \\
Ours w/o $\mathbb{B}$ &  1.63 & 293.6 \\
Ours w/o $\mathbb{C}$ &  1.69 & 291.8 \\
Ours w/o $\mathbb{D}$ &  1.62 & 295.5 \\
\bottomrule
\end{tabular}
\end{table}

\begin{figure}
    \centering
    \includegraphics[width=0.95\linewidth]{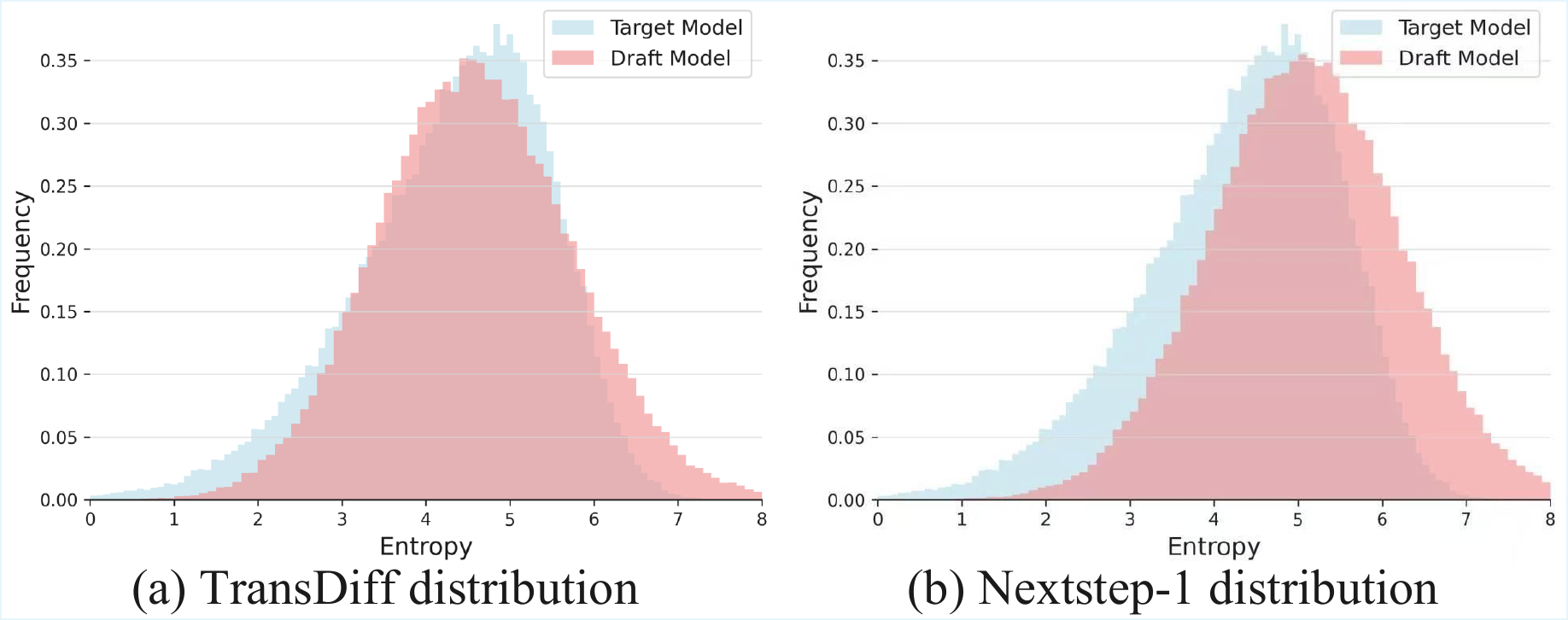}
    \caption{Entropy distribution alignment between the draft model (after entropy-informed training) and the target model.
    }
    \vspace{-1em}
    \label{last_entropy}
\end{figure}

\section{Discussion}
Continuous-space AR+Diffusion avoids quantization loss to enable high-fidelity generation, but its continuity creates unique entropy challenges. Unlike NLP AR or discrete visual AR, it lacks built-in constraints to regulate entropy, making explicit entropy guidance critical for acceleration.   The high information density of language naturally preserves diversity: small draft models align with large ones in entropy without extra guidance.  Discrete visual tokens  reduce visual complexity into finite primitives, but their entropy is bounded by codebook size.  While quantization loss introduces artifacts, the discrete nature ensures draft models can learn token co-occurrence patterns (e.g., ``sky tokens" adjacent to ``cloud tokens") with reasonable entropy alignment .  Our entropy loss targets this: by penalizing low-entropy feature distributions, it forces draft models to retain appropriate alignment.

In Figure~\ref{last_entropy}, we show the entropy distribution of the draft model after entropy-informed training, which is closely overlapped with the target model—mirroring the consistency of entropy seen in language models. This alignment reduces the rejection rates of the draft model by over 40\%, as the draft’s feature diversity now matches the target’s.

\section{Conclusion}
This work addresses the critical inference latency bottleneck of AR+Diffusion hybrid paradigms, which excel in structured semantic modeling and high-fidelity generation, but struggle with real-time deployment. We propose a unified acceleration framework: entropy-informed speculative decoding resolves the entropy mismatch of AR draft models with target models to boost hit rates; distribution matching distillation with initialization adaptation stabilizes diffusion distillation and achieves extreme denoising step reduction; and an end-to-end training pipeline with early entropy monitoring balances speed and quality. Experiments validate that our scheme achieves nice inference speedup, with minimal performance degradation. This framework provides a paradigm for the synergistic acceleration of hybrid generative models, bridging the gap between performance and  deployment.

%% file: sec/3_preliminary_xiaoxiao.tex
\section{Preliminary}

\subsection{Image Generation Paradigm}
In image generation, AR models replace the linguistic tokens of NLP with ``visual elements" (pixels, latent features, or discrete tokens). The joint distribution of an image can thus be factorized as:
\begin{equation}
    p(\mathbf{x}) = \prod_{i=1}^{N} p(x_i \mid x_1, x_2, ..., x_{i-1}),
\end{equation}
where \(\mathbf{x}\) is the image, \(x_i\) is the $i$-th visual element, and $N$ is the total number of elements. Modern visual AR models employ discrete tokenization—converting images into compact discrete token sequences via specialized image tokenizers, then allowing efficient autoregressive modeling that balances quality and scalability. This design is exemplified by the architecture and experiments of LlamaGen~\cite{sun2024llamagen}.

Diffusion models, in contrast, are defined by a two-stage probabilistic process.  
The forward diffusion process is a Markov chain that progressively corrupts the original data \(x_0\) (e.g., images or videos) by incrementally adding Gaussian noise until the data is transformed into a pure Gaussian noise distribution \(x_T\):
\begin{equation}
    q(x_t \mid x_{t-1}) = \mathcal{N}(x_t; \sqrt{1-\beta_t}x_{t-1}, \beta_t \mathbf{I}),
\end{equation}
where \(\beta_t \in (0,1)\) denotes the time-dependent noise schedule and \(\mathbf{I}\) is the identity matrix.  
The reverse denoising process learns to invert forward corruption by training a parameterized neural network to approximate the conditional distribution \(p_\theta(x_{t-1} \mid x_t)\) and progressively reconstruct clean data from noise.

\subsection{AR+Diffusion Hybrid Paradigm}
The AR+Diffusion hybrid paradigm fuses AR and diffusion models to address the inherent limitations of each individual paradigm: pure AR suffers from sequential latency and quantization error, while pure diffusion lacks structured semantic guidance. Its core is leveraging AR’s hierarchical semantic modeling with diffusion’s high-fidelity synthesis.

MAR~\cite{li2024mar} pioneers this hybrid approach by discarding discrete tokenization and operating directly on continuous latents\(x \in \mathbb{R}^{d}\) while preserving AR’s sequential masked prediction. It introduces a diffusion-inspired objective:
\begin{equation}
\mathcal{L}_{MAR} = \mathbb{E}_{\epsilon \sim \mathcal{N}(0,I), t \sim \text{Uniform}(0,1)}\left[\|\epsilon - \epsilon_\theta(x_t \mid t, z_{AR})\|^2\right],
\end{equation}
where \(x_t = \sqrt{\bar{\alpha}_t}x + \sqrt{1-\bar{\alpha}_t}\epsilon\) is the noisy latent, \(z_{AR}\) denotes AR’s predicted feature, and \(\epsilon_\theta\) represents the diffusion noise estimator. 

TransDiff~\cite{zhen2025marrying} further enables end-to-end integration: its AR transformer models sequential semantic features as \(z = ART(x_{1:i-1}, c)\), where $c$ is class/text condition and \(x_{1:i-1}\) are preceding latents. A DiT-based diffusion decoder then reconstructs images via: \(x_0 = \text{Diff}_\theta(x_t \mid t, z)\). Although TransDiff is capable of generating all tokens in one step, we treat it as a multi-step method for improved performance.

\subsection{Acceleration}
Speculative decoding is a widely adopted acceleration technique for AR generative models. It mitigates the latency caused by sequential token generation through a ``draft–verify" mechanism, where a lightweight draft model proposes multiple tokens in parallel and a stronger target model verifies them.
EAGLE (Extrapolation Algorithm for Greater Language-model Efficiency)~\cite{li2024eagle} and its successors~\cite{li2024eagle2,li2025eagle} represent the state-of-the-art in speculative decoding. 
They address key limitations of vanilla speculative sampling by introducing feature-level modeling and maintaining consistency between training and inference. Specifically, EAGLE employs both feature regression and classification objectives:
\begin{equation}
\label{eq_loss_reg_cls}
\begin{aligned}
     \mathcal{L}_{\text{reg}} &= \text{Smooth L1}(f_{j+1}, \text{Draft Model}(T_{2:j+1}, F_{1:j})),
     \\
     \mathcal{L}_{\text{cls}} &= \text{Cross-Entropy}(p_{j+2}, \hat{p}_{j+2}),
\end{aligned}
\end{equation}
where $F_{1:j}$ denotes draft model's current feature sequence, $\hat{f}_{j+1}$ the target model’s next-step feature, $T_{2:j+1}$ the output tokens, and $p/\hat{p}$ their discrete probability distributions.

For diffusion models, the acceleration faces a distinct challenge: generating high-quality samples typically requires hundreds to thousands of denoising steps. DMD~\cite{yin2024one} aims to distill multi-step diffusion models into efficient one-step generators by minimizing the reverse Kullback-Leibler (KL) divergence between the generated distribution \(p_{\text{fake}}\) from the one-step student generator \(G_\theta\) and the target distribution \(p_{\text{real}}\) from the pre-trained teacher diffusion model.

%% file: sec/X_suppl.tex
\clearpage
\setcounter{page}{1}
\maketitlesupplementary

\begin{table*}[htbp]
\centering
\caption{TransDiff Model Configurations (Target Models: L/H Variants; Draft Model: Unified 6-Layer Design)}
\label{tab:transdiff_config_en}
\resizebox{\linewidth}{!}{ 
\begin{tabular}{l l c c c c c c} 
\hline
\multirow{2}{*}{\textbf{Model Variant}} & \multirow{2}{*}{\textbf{Component Type}} & \multicolumn{3}{c}{\textbf{Encoder Parameters}} & \multicolumn{3}{c}{\textbf{Decoder Parameters}}  \\
\cline{3-8} 
& & \textbf{Embed Dim} & \textbf{Depth (Layers)} & \textbf{Num Heads} & \textbf{Embed Dim} & \textbf{Depth (Layers)} & \textbf{Num Heads} \\ 
\hline
\multirow{2}{*}{\textbf{TransDiff-L}} 
& Target AR & 1024 & 16 & 16 & 1024 & 16 & 16 \\ 
& Draft AR  & 1024  & 6  & 16  & -    & -  & - \\ 
\hline
\multirow{2}{*}{\textbf{TransDiff-H}} 
& Target AR & 1280 & 20 & 16 & 1280 & 20 & 16 \\ 
& Draft AR  & 1280  & 6  & 16  & -    & -  & - \\ 
\hline
\multicolumn{8}{l}{\small \textit{Note:} The Draft AR only participates in autoregressive speculative decoding and does not include an independent decoder module (decoder parameters marked as ``-");} \\
\multicolumn{8}{l}{\small The embed dimension and number of heads of the Draft AR are set the same with the corresponding Target AR to maintain consistency.} 
\end{tabular}
}
\end{table*}

\begin{table*}[h]
\centering
\caption{Complete Class-conditional generation on ImageNet 256×256.}
\label{tab:complete_imagenet_gen}
\resizebox{0.99\linewidth}{!}{
\begin{tabular}{lccccccc}
\toprule
\textbf{Method} & \textbf{Latency/s}& \textbf{Speedup}  & \textbf{FID ($\downarrow$)} &\textbf{sFID ($\downarrow$)} & \textbf{IS ($\uparrow$)}& \textbf{Precision ($\uparrow$)} & \textbf{Recall ($\uparrow$)}\\
\midrule
\multicolumn{4}{l}{\textbf{\textit{MAR}}} \\

\midrule
MAR-L & 5.31 & 1.00 & 1.78 &3.27&296.0 &0.81&0.60\\
MAR-L + DMD&  1.99 & 2.67 & 1.81 &3.51&295.5&0.78&0.59 \\
MAR-L + LazyMAR&  2.29 & 2.32 & 1.93 &3.43&297.4&0.80&0.60 \\
MAR-L + Ours  & 1.21 & 4.38  & \textbf{1.77} &3.31&\textbf{297.1} &0.81&0.60\\
MAR-H & 9.97&1.00  & 1.55 &3.09& 303.7 &0.81&0.62\\
MAR-H + DMD& 4.17  & 2.39& 1.73 &3.46& 301.0 &0.81&0.61\\
MAR-H + LazyMAR& 4.24 &2.35 &  1.69 &3.13& 299.2 &0.80&0.61\\
MAR-H + Ours & 2.23 &4.47 &  \textbf{1.61} &3.10& \textbf{304.3}&0.81&0.62 \\
\midrule
\multicolumn{4}{l}{\textbf{\textit{TransDiff}}} \\

\midrule
TransDiff-L & 3.17 & 1.00 & 1.61 &3.03& 295.1 &0.81&0.60\\
TransDiff-L + SD  &1.75 &1.81 & 1.88 &3.59& 283.6 &0.80&0.60\\
TransDiff-L + DMD  & 1.61&1.97 & 1.79 &3.30& 288.3 &0.81&0.60\\
TransDiff-L + Ours  &0.68 & {4.69} & \textbf{1.66} &3.11& \textbf{294.1} &0.81&0.60\\
TransDiff-H & 6.72 & 1.00 & 1.55 &2.93& 297.9 &0.81&0.61\\
TransDiff-H + SD & 3.52  & 1.91 & 1.68 &3.17& 291.1 &0.80&0.60\\
TransDiff-H + DMD & 3.48 & 1.93 & 1.71 &3.09&  289.9 &0.80&0.61\\
TransDiff-H + Ours & 1.38 & 4.88 & \textbf{1.60} &2.99& \textbf{295.0} &0.81&0.61\\
\bottomrule
\end{tabular}
}
\end{table*}

\section*{S1. Introduction to Supplementary Material}
This supplementary material complements the main text by providing \textbf{detailed implementation details}, \textbf{extended experimental results}, \textbf{continuous speculative decoding algorithm}, and \textbf{visual results} for the proposed Fast-ARDiff framework. All content focuses on information critical to evaluating method validity and enabling independent reproduction. No new contributions are introduced but only expansions of content constrained by the main text’s length.

\section*{S2. Model Architecture \& Implementation Details}
\subsection*{S2.1 AR Branch Detailed Design}
The AR branch consists of a \textit{Target AR} (for high-quality semantic guidance) and a \textit{Draft AR} (for speculative decoding), both based on the Transformer encoder architecture. Table~\ref{tab:transdiff_config_en} shows the details of them. We take TransDiff-L for example in the following.

\noindent \textbf{Entropy Calculation Logic: }Entropy is computed on the \textit{penultimate layer} attention maps (Layer 15 for Target AR, Layer 5 for Draft AR) after softmax normalization. For an attention map \(A \in \mathbb{R}^{R \times R}\), the entropy loss is defined as:
\begin{equation}
L_{entropy} = \frac{1}{R \times N} \sum_{r=1}^{R} \sum_{c=1}^{N} \Omega_{r,c} \log(\Omega_{r,c})
\end{equation}
where \(N\) denotes the number of attention heads in Draft AR. This loss penalizes low-entropy (overconfident) distributions by minimizing negative entropy.


\subsection*{S2.2 Diffusion Branch Detailed Design}
Diffusion distillation (CD + DMD stages) does \textbf{not} alter the model size. The diffusion branch retains an identical architecture.

\subsection*{S2.3 Training Details}
In the experiments on the ImageNet dataset, the training process is divided into two sequential stages with detailed configurations as follows:

\begin{enumerate}[leftmargin=*]
    \item \textbf{Stage 1 (Consistency Distillation, 20 Epochs)}
    \begin{itemize}[leftmargin=1.5em]
        \item \textit{Optimizer}: Adam optimizer with momentum parameters \(\epsilon_1 = 0.9\) and \(\epsilon_2 = 0.95\).
        \item \textit{Training Hyperparameters}: Batch size = 32; initial learning rate = \(1 \times 10^{-4}\).
        \item \textit{Guidance Strategy}: Classifier-Free Guidance (CFG) with a scale of 1.75 is applied across all time steps.
    \end{itemize}

    \item \textbf{Stage 2 (Distribution Matching Distillation, 10 Epochs)}
    \begin{itemize}[leftmargin=1.5em]
        \item \textit{Training Hyperparameters}: Batch size = 32; generator learning rate = \(2 \times 10^{-5}\); learning rate of the fake branch = \(1 \times 10^{-5}\).
        \item \textit{Guidance Strategy}: CFG is already distilled in stage 1, so in this stage we use fixed CFG embedding.
    \end{itemize}
\end{enumerate}

\section*{S3. Inference for Continuous-Space Autoregressive Speculative Decoding}

Continuous-Space Speculative Decoding (CSpD)~\cite{wang2024continuous} is the first framework to extend speculative decoding from discrete token spaces to continuous-valued AR image generation, addressing the unique challenges of continuous probability density functions (PDFs) and intractable modified distributions. Its core innovations include three key components: 

\noindent\textbf{1.Denoising Trajectory Alignment:} By sharing the same Gaussian noise \(\epsilon_t \sim \mathcal{N}(0,I)\) at each denoising step, CSpD aligns the output distributions of draft and target models—mitigating trajectory divergence that causes low acceptance rates. This alignment also simplifies the calculation of the continuous PDF ratio \(\frac{p(x)}{q(x)}\) via the product of variance ratios along denoising steps. 

\noindent\textbf{2.Continuous Acceptance Criterion:} Leveraging the diffusion process inherent to continuous AR models (e.g., MAR), the PDF ratio \(\frac{p(x)}{q(x)}\) is computed as \(\Sigma \cdot \frac{p_{\theta}(x_0|x_1^p)}{q_{\theta}(x_0|x_1^q)}\), where \(\Sigma = \prod_{t=2}^T \sqrt{\frac{|\Sigma_{q,t}|}{|\Sigma_{p,t}|}}\) ( \(\Sigma_{q,t}/\Sigma_{p,t}\) denote draft/target variances at step \(t\) ). 

\noindent\textbf{3.Integral-Free Modified Distribution Sampling:} For rejected draft tokens, CSpD uses acceptance-rejection sampling with a properly defined upper bound \(M=1/Z\) ( \(Z\) is the intractable integral of \(max(0,p(x)-q(x))\) ), avoiding complex integration while preserving the target distribution.

Notably, CSpD requires no additional training, model architecture modifications, or changes to the target output distribution—enabling seamless integration with existing continuous AR models (e.g., MAR-B/L/H) and achieving up to 2.33× speedup on ImageNet 256×256 generation. However, vanilla CSpD still suffers from redundant target model verification when drafts generate low-diversity (low-entropy) features (e.g., homogeneous backgrounds), as these features often lead to unnecessary acceptance checks. 

To address this, we integrate an entropy verification mechanism into CSpD in Algorithm~\ref{alg:cspd_entropy_infer}: during draft token generation, we monitor the entropy of shallow-layer attention maps from the draft model. Low entropy (signaling overconfident, low-diversity features) triggers immediate speculation termination, avoiding redundant verification and further improving inference efficiency, while retaining all advantages of vanilla CSpD.

\begin{algorithm*}[htbp]
   \caption{CSpD Inference with Entropy Verification Mechanism}
   \label{alg:cspd_entropy_infer}
\begin{algorithmic}[1]
        \State \textbf{Input:} Target AR ($M_p$), Draft AR ($M_q$), Prefix Tokens (pre-filled with $\eta\%$ target tokens), Draft Length $\gamma$, Denoising Steps $T$, Temperature $\tau$, CFG Scale $\sigma$, Entropy Threshold $\tau_{\text{ent}}$
        \State \textbf{Initialize:} $\text{accepted\_tokens} \leftarrow \text{Prefix Tokens}$, $\epsilon_t \sim \mathcal{N}(0, I)$ (shared for trajectory alignment), $\text{speculate\_flag} \leftarrow \text{True}$, $\text{draft\_tokens} \leftarrow []$
        
        \State \textbf{Step 1: Generate Draft Tokens with Entropy Verification}
        \For {$i := 1$ to $\gamma$}
            \If {$\text{speculate\_flag} = \text{True}$}
                \State $q_i(x) \leftarrow M_q(\text{accepted\_tokens} + \text{draft\_tokens}, \sigma, \tau)$  
                \State $x_i \sim q_i(x)$  
                \State $\text{draft\_tokens} \leftarrow \text{draft\_tokens} + [x_i]$
                \State $\Omega^s \leftarrow \text{Softmax}(M_q.\text{shallow\_attn}(\text{accepted\_tokens} + \text{draft\_tokens}), \text{dim}=-1)$  
                \State $\mathcal{H}^s \leftarrow -L_{\text{entropy}}$  
                \If {$\mathcal{H}^s < \tau_{\text{ent}}$}
                    \State $\text{speculate\_flag} \leftarrow \text{False}$
                    \State $\text{draft\_tokens} \leftarrow \text{draft\_tokens}[1:i-1]$  
                    \State \textbf{break}
                \EndIf 
            \Else
                \State \textbf{break}
            \EndIf
        \EndFor 
        
        \State \textbf{Step 2: Parallel Verification by Target AR}
        \For {$i := 1$ to $\text{len}(\text{draft\_tokens}) + 1$}
            \State $p_i(x) \leftarrow M_p(\text{accepted\_tokens} + \text{draft\_tokens}[1:i-1], \sigma, \tau, \epsilon_t)$  
        \EndFor
        
        \State \textbf{Step 3: Compute Acceptance Criterion Factor $\Sigma$}
        \State $\Sigma \leftarrow \prod_{t=2}^{T} \sqrt{\frac{|\Sigma_{q,t}|}{|\Sigma_{p,t}|}}$  
        
        \State $\frac{p_i(x)}{q_i(x)} \leftarrow \Sigma \cdot \frac{p_i(x \mid x_1^p)}{q_i(x \mid x_1^q)}$  
        
        \State \textbf{Step 4: Determine Accepted Draft Tokens}
        
        \State $r_1, \dots, r_{\text{len}(\text{draft\_tokens})} \sim U(0, 1)$  
        \State $n \leftarrow \min\left( \{i-1 \mid 1 \leq i \leq \text{len}(\text{draft\_tokens}), r_i > \frac{p_i(x)}{q_i(x)}\} \cup \{\text{len}(\text{draft\_tokens})\} \right)$  
        \State $\text{accepted\_tokens} \leftarrow \text{accepted\_tokens} + \text{draft\_tokens}[1:n]$  
        
        \State \textbf{Step 5: Resample for Rejected Tokens (if $n < \text{len}(\text{draft\_tokens})$)}
        \If {$n < \text{len}(\text{draft\_tokens})$}
            
            \Repeat
                \State $x_t \sim p_{n+1}(x \mid x_1^p)$  
                \State $\alpha_s \leftarrow \frac{\max(0, \Sigma \cdot p_{n+1}(x_t \mid x_1^p) - q_{n+1}(x_t \mid x_1^q))}{\Sigma \cdot p_{n+1}(x_t \mid x_1^p)}$  
                \State $\epsilon \sim U(0, 1)$
            \Until {$\epsilon \leq \alpha_s$}  
            \State $\text{accepted\_tokens} \leftarrow \text{accepted\_tokens} + [x_t]$  
        \EndIf
        
        \State \textbf{Output:} Updated Token Sequence $\text{accepted\_tokens}$  
    \end{algorithmic}
\end{algorithm*}

\section*{S4. Extended Experimental Results}
We present the complete results on ImageNet in Table~\ref{tab:complete_imagenet_gen}, while the main text only shows the core parts due to space constraints.

\section*{S5. More visual results}
\begin{figure*}[htbp]
    \centering
    \includegraphics[width=1 \linewidth]{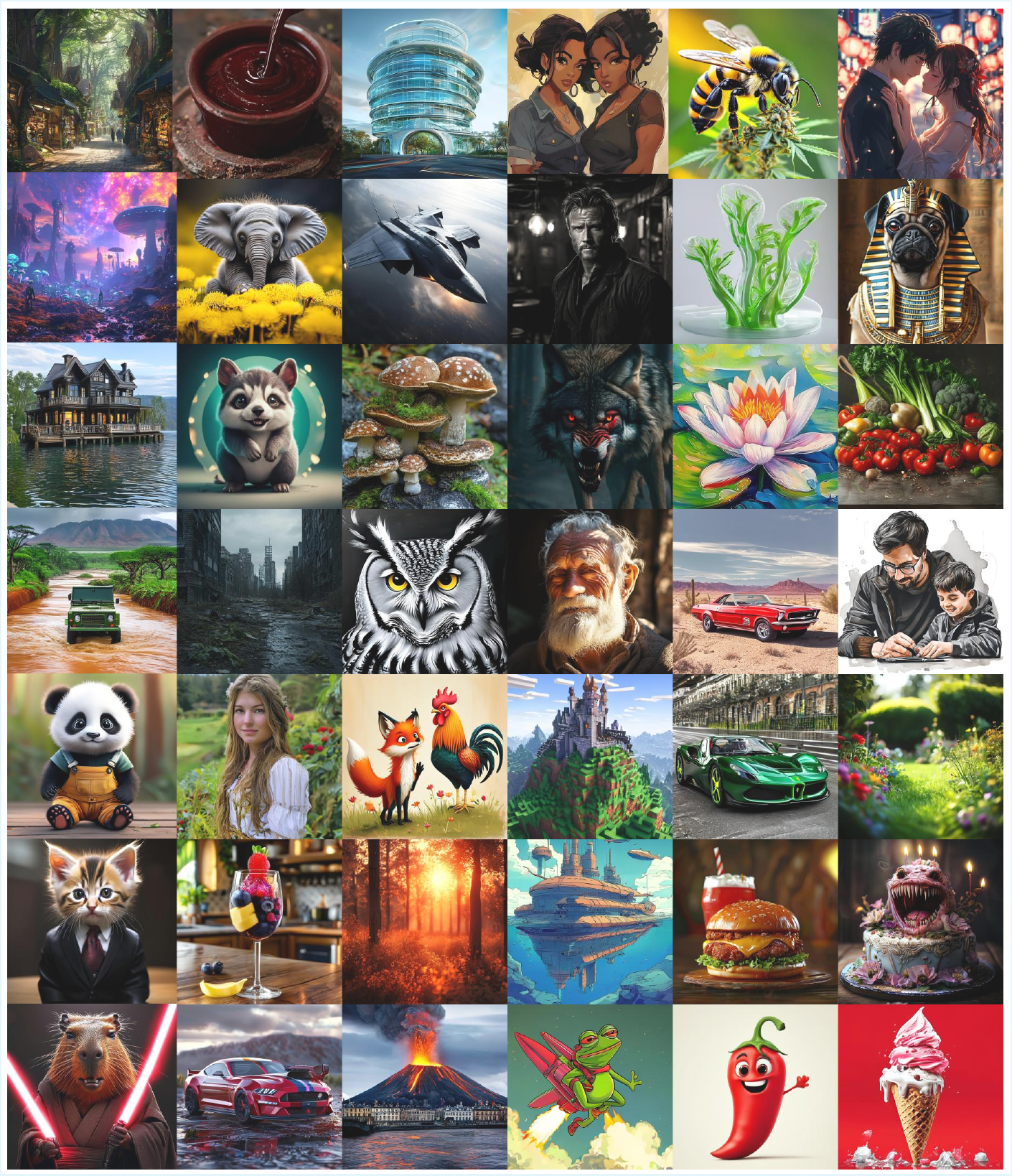}
    \caption{More visual results on MJHQ-30K with NextStep-1 .
    }
    \label{mjhq_show_sup}
    \vspace{-1em}
\end{figure*}

\begin{figure*}[htbp]
    \centering
    \includegraphics[width=1 \linewidth]{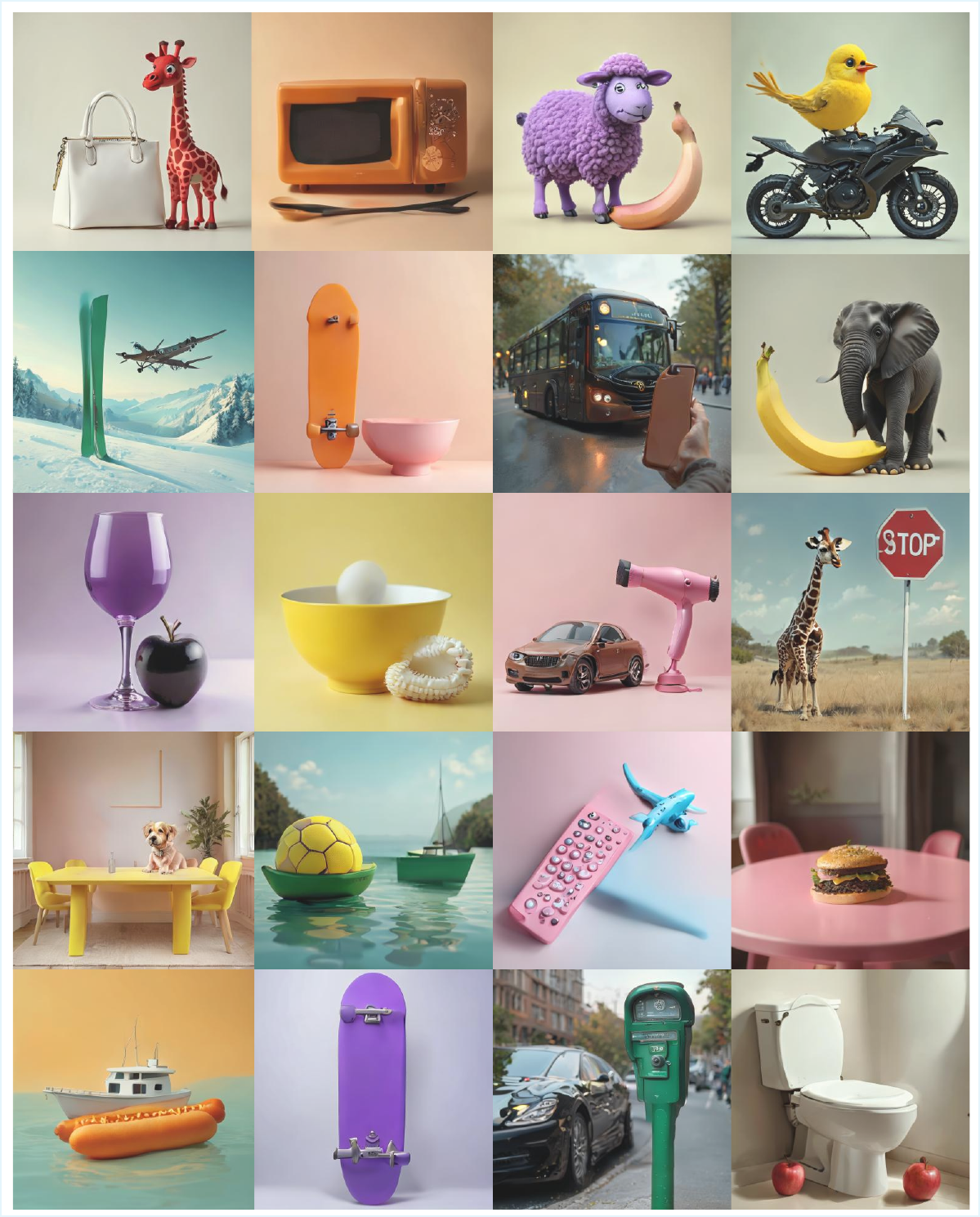}
    \caption{More visual results on GenEval with NextStep-1.
    }
    \label{geneval_show_sup}
    \vspace{-1em}
\end{figure*}

\begin{figure*}[htbp]
    \centering
    \includegraphics[width=1 \linewidth]{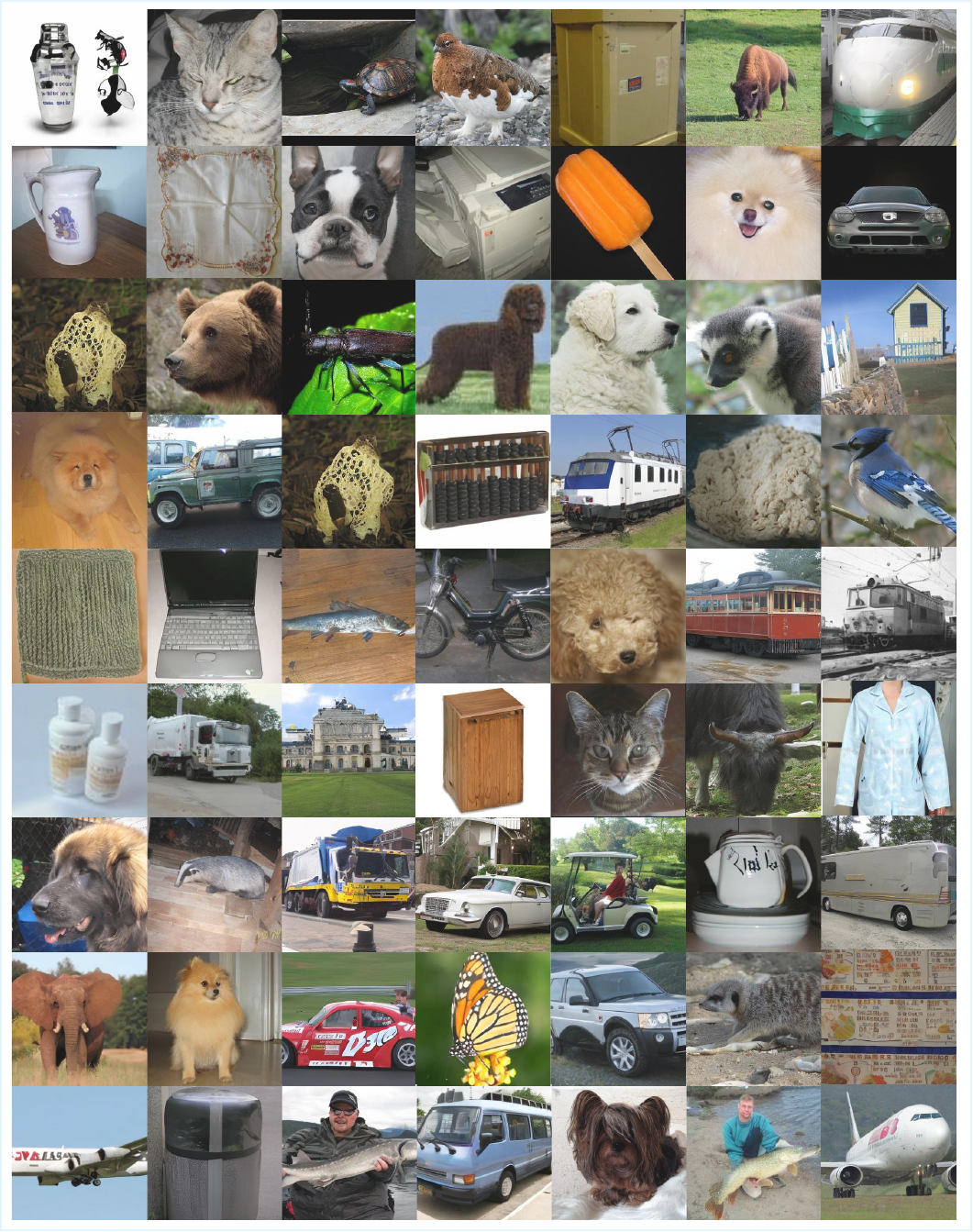}
    \caption{More visual results on ImageNet 256 $\times$ 256 with TransDiff-L.
    }
    \label{imagenet_show_sup}
    \vspace{-1em}
\end{figure*}

We provide additional qualitative results on several datasets in Figures~\ref{mjhq_show_sup}, \ref{geneval_show_sup}, and \ref{imagenet_show_sup}, which demonstrate that our method achieves generation quality.

